# Using Modified Partitioning Around Medoids Clustering Technique in Mobile Network Planning

Lamiaa Fattouh Ibrahim[1, 2]   Manal Hamed Al Harbi[3]

[1]Department of Computer Sciences and Information System, Institute of Statistical Studies and Research, Cairo University Giza, Egypt
Currently
[2] Department of Computer Sciences, Faculty of Computer and Information Technology King AbdulAziz University
Jeddah, Saudi Arabia
[3]College of Education, UMM AL-QURA University, Macca, Saudi Arabia

lfattouh@mailer.eun.eg

## Abstract

Every cellular network deployment requires planning and optimization in order to provide adequate coverage, capacity, and quality of service (QoS). Optimization mobile radio network planning is a very complex task, as many aspects must be taken into account. With the rapid development in mobile network we need effective network planning tool to satisfy the need of customers. However, deciding upon the optimum placement for the base stations (BS's) to achieve best services while reducing the cost is a complex task requiring vast computational resource. This paper introduces the spatial clustering to solve the Mobile Networking Planning problem. It addresses antenna placement problem or the cell planning problem, involves locating and configuring infrastructure for mobile networks by modified the original Partitioning Around Medoids PAM algorithm. M-PAM (Modified Partitioning Around Medoids) has been proposed to satisfy the requirements and constraints. PAM needs to specify number of clusters (k) before starting to search for the best locations of base stations. The M-PAM algorithm uses the radio network planning to determine k. We calculate for each cluster its coverage and capacity and determine if they satisfy the mobile requirements, if not we will increase (k) and reapply algorithms depending on two methods for clustering. Implementation of this algorithm to a real case study is presented. Experimental results and analysis indicate that the M-PAM algorithm when applying method two is effective in case of heavy load distribution, and leads to minimum number of base stations, which directly affected onto the cost of planning the network.

## 1. Introduction

The network planning process has to consider a variety of constraints including: policy of administrations, planning objective, etc, there is no universal method that is applicable to all network planning problems. Due to the complexity of this process artificial intelligence (AI) [1], clustering techniques [2] - [5], Ant-Colony-Based algorithm [2], [6] have been successfully deployed in wire network planning. Tabu Search TS [7], [8], heuristic algorithm [9] and genetic algorithm (GA) [10] have been successfully deployed in mobile network planning.

Cell planning in GSM mobile system is one of the most important operations must be done before the installation of the system, cell planning means studying the geographic area where the system will be installed and the radius of each BT coverage and the frequencies used [11]. Cellular telephony is designed to provide communications between two moving units, called mobile stations (MS's), or between one mobile unit and one stationary unit, often called a land unit [12]. A service provider must be able to locate and track a caller, assign a channel to the call, and transfer the channel from base station to base station as the caller moves out of range. Each cellular service area is divided into regions called cells. Each cell contains an antenna and is controlled by a solar or AC power network station, called the base station (BS). Each base station, in turn, is controlled by a switching office, called a mobile switching center (MSC). The MSC coordinates communication between all the base stations and telephone central office. Cell planning is challenging due to inherent complexity, which stems from requirements concerning radio modeling and optimization. Manual human design alone is of limited use in creating highly optimized networks, and it is imperative that intelligent computerized technology is used to create appropriate network designs [13].

Data mining is an expanding area of research in artificial intelligence and information management. The objective of data mining is to extract relevant information from databases containing large amounts of information.

Typical data mining and analysis tasks include classification, regression, and clustering of data, determining parameter dependencies, and finding various anomalies from data [1].

Clustering analysis is a sub-field in data mining that specializes in techniques for finding similar groups in large database [14]. Its objective is to assign to the same cluster data that are more close (similar) to each other than they are to data of different clusters. The application of clustering in spatial databases presents important characteristics. Spatial databases usually contain very large numbers of points. Thus, algorithms for clustering in spatial databases do not assume that the entire database can be held in main memory. Therefore, additionally to the good quality of clustering, their scalability to the size of the database is of the same importance [15]. In spatial databases, objects are characterized by their position in the Euclidean space and, naturally, dissimilarity between two objects is defined by their Euclidean distance [16].

This paper introduces the spatial clustering to solve the Mobile Networking Planning problem. Section 2 discusses main phases used in radio network planning. In sections 3 The Cluster Partitioning Around Medoids (PAM) are reviewed. In section 4, the proposed algorithm is fully described. A case study is presented in section 5. Section 6, the Comparison between Proposed Methods and Other Methods. The paper conclusion and Future Work is presented in section 7.

## 2. Main Phases Used in Radio Network Planning

The radio network planning process can be divided into different phases [17]. At the beginning is the Preplanning phase. In this phase, the basic general properties of the future network are investigated, for example, what kind of mobile services will be offered by the network, what kind of requirements the different services impose on the network, the basic network configuration parameters and so on. The second phase is the main phase. A site survey is done about the to-be-covered area, and the possible sites to set up the base stations are investigated. All the data related to the geographical properties and the estimated traffic volumes at different points of the area will be incorporated into a digital map, which consists of different pixels, each of which records all the information about this point. Based on the propagation model, the link budget is calculated, which will help to define the cell range and coverage threshold. There are some important parameters which greatly influence the link budget, for example, the sensitivity and antenna gain of the mobile equipment and the base station, the cable loss, the fade margin etc. Based on the digital map and the link budget, computer simulations will evaluate the different possibilities to build up the radio network part by using some optimization algorithms. The goal is to achieve as much coverage as possible with the optimal capacity, while reducing the costs also as much as possible. The coverage and the capacity planning are of essential importance in the whole radio network planning. The coverage planning determines the service range, and the capacity planning determines the number of to-be-used base stations and their respective capacities.

In the third phase, constant adjustment will be made to improve the network planning. Through driving tests the simulated results will be examined and refined until the best compromise between all of the facts is achieved. Then the final radio plan is ready to be deployed in the area to be covered and served. The whole process is illustrated as the Fig. 1.

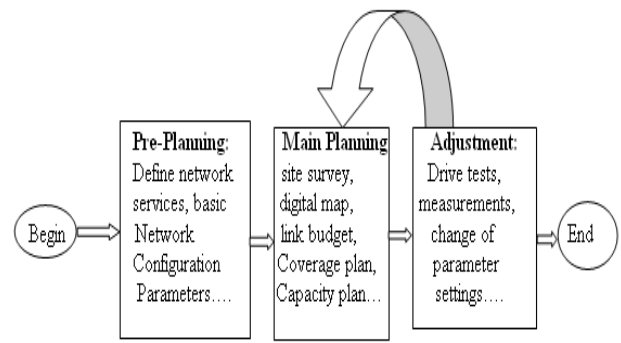

Fig. 1 Radio network planning process

The two important mobile technologies are: GSM Global System for Mobile Communications and UMTS Universal Mobile Telecommunications System [17]. This paper use GSM technology.

GSM referred to as 2G. It operates in the frequency 900-Mhz and a variation of it operates in the 1800-Mhz. GSM planning divided into two phases, coverage planning phase, capacity planning phase. The coverage planning and capacity planning are independent. The frequency is a one of the important issue resource in GSM system.

2.1 Coverage Planning In GSM

The coverage planning depends on the received signal strength. Base stations are placed to ensure that the signal strength is sufficiently high in all areas of the region to be served. In this stage the link budget and Okumura-Hata function are calculated, which will help to define the cell range.

When considering the coverage of a cell, the maximum radius of the cell must be determined. Coverage is determined with respect to the maximum path loss that can be applied to the signal. The maximum path loss is calculated for the reverse link since the transmission power of subscriber antenna is much less than that of the base station. Link budget is designed to calculate the maximum path loss. It is defined in [18] as: the accounting of all of the gains and losses from the radio transmitter (source of the radio signal), through cables, connectors and free air to the receiver. A simple link budget equation looks like this:

Allowed propagation loss = Transmitted EIRP + Receiver Gains − Total margin (Losses)

### 2.1.1 The Elements of a Link Budget

The elements can be broken down into three main parts:
1. Transmitting side with effective transmit power.
2. Receiving side with effective receiving sensibility
3. Propagation part with propagation losses (total margins).

A complete radio link budget is simply the sum of all contributions (in decibels) across the three main parts of a transmission path. All positive values are gain and all negative values are losses. The introduction of these elements will be presented in the following section.

#### 2.1.1.1 Transmitting side with effective transmit power

It can be calculated by the following equation:
 Effective Transmit Power [EIRP] = Transmitter power − (Cable TX loss + Body TX loss) + Antenna TX gain
Where:
Transmit power (Tx): The transmit power is the power output of the radio card. The transmit power of card can normally be found in the vendor's technical specification.
Cable Loss: Losses in the radio signal will take place in the cables that connect the transmitter and the receiver to the antennas. The losses depend on the type of cable and frequency of operation.
Body Loss: Allow at least 0.25 dB (loss) for each connector in cabling.
Antenna Gain: is defined as the ratio of the radiation intensity of an antenna in a given direction, to the intensity of the same antenna as it radiates in all directions (isotropically).

#### 2.1.1.2 Receiving side with effective receiving sensibility

It can be calculated by the following equation:
 Effective Receiving Sensibility = Receiver Sensibility − (Cable RX loss + Body RX loss) + Antenna RX gain
Where:
cable loss , body loss and antenna gain like transmitter side above.
Receiver Sensibility: The sensibility of a receiver is a parameter that deserves special attention as it indicates the minimum value of power that is needed to successfully decode/extract "logical bits" and achieve a certain bit rate.

#### 2.1.1.3 Propagation part with propagation losses

The propagation losses are related to all attenuation of the signal that takes place when the signal has left the transmitting antenna until it reaches the receiving antenna. One of the main causes for the power of a radio signal to be lost in the air is fading. Shadow fading is a phenomenon that occurs when a mobile moves behind an obstruction and experiences a significant reduction in signal power.
   Total Margins = Fading Margin + Interference Margin + Penetration Margin + Other Margins

Once the maximum allowed propagation loss in cell is known, the maximum cell range and coverage area can be evaluated by applying a model like Okumura-Hata model for propagation loss [18]. Propagation model is the algorithm that the predicate tool uses to calculate signal strength. Each model is developed to predicate propagation in particular environments such as overlay, open area, suburban area, urban area, high dense urban and low dense urban. Okumura-Hata model is widely used for coverage calculation in macrocell network planning (taking from lesson in RF- Basic Concept: Technical Parameter & Link Budget, CISCOM Cellular Integrated Services Company).

The Okumura-Hata model is valid for the following conditions:
- Environment is urban, suburban or open area
- Frequency is in the range 150-1000 MHz (recommended).
- Antenna height of the base station is in the range 30-200 meters (recommended).
- Antenna height of the mobile station is in the range 1-10 meters (recommended).
- Distance between the base station and mobile station is in the range 1-20 km (recommended).

The path loss is expressed as the sum $A + B \log10(d) + C$, where the constant coefficients A, B, and C are dependent upon the propagation terrain, and d is the distance between the transmitter and receiver.

The parameters A and B are set by the user according to Table 1 (taking from Alcatel GSM Network). These values have been determined by fitting the model with measurements.

Table 1 Parameters A and B in GSM

| Frequency | 900 MHz | 1800 MHz |
|---|---|---|
| A | 69.55 | 46.3 |
| B | 26.16 | 33.9 |

### 2.2 Capacity Planning In GSM

The Erlang (E) is a unit of measurement of traffic intensity. It can be calculated by this equation:

A= n * T/ 3600  Erlang

Where: A= offered traffic from one or more users in the system, n = number of calls per hour and T=average time call in seconds.

Capacity planning depends mainly on the frequency allocation. In this stage we calculate maximum cell capacity.  System capacity planning is divided into two parts:

1. The first part is to estimate a signal transceiver and site capacity. Required parameters are: number of subscribers, available frequencies, number of cells per pattern, number of channels per one carrier frequency, cell pattern, grade of service, number of calls per hour, average call time in seconds, control channels.
2. The second part of the process is to estimate how many mobile users each cell can serve. Once the cell capacity and subscriber traffic profiles are known, network area base station requirements can be calculated. Estimations can be done in Erlangs per subscriber or kilobits per subscriber.

Traffic per subscribers (TSUB) = Average call time * Number of calls / 3600

Frequencies per cell = Available frequencies/Number of cells per pattern

Traffic channels per cell = Frequencies per cell * Number of  channels per one carrier frequency  – control channels

Traffic per cell = (From Erlang table), Traffic channels per cell With GOS  implies Traffic per cell

The number of subscribers per cell =  Traffic per cell / Traffic per subscribers

The number for cells needed = Number of subscribers/Number of subscribers per cell

## 3. The Cluster Partitioning Around Medoids (PAM)

Spatial clustering algorithms can be classified into four categories. They are the partition based, the hierarchical based, the density based and the grid based [19], [20] as Fig. 2. Among all the clustering methods, we found that the partitioning based algorithm to be most suitable since our objective is to discover good locations that are hidden in the data. Partitioning based clustering methods include two major categories, k-means and k- medoids. The common premise of these two methods is to randomly partition the database into k subsets and refine the cluster centers repeatedly to reduce the cost function. The cost function in the spatial domain is the sum of distance error distance *E* from all data objects to their assigned centers.

The non center data points are assigned to the centers that they are nearest to it. The k-means algorithm is one of the first clustering algorithms proposed. It easy to understand and implement, and also known for its quick termination. The k-means algorithm defines the cluster centers to be the gravity center of all the data points in the same cluster. In regular planar space, the cluster gravity center guarantees the minimum sum of distances between the cluster members and itself. However, the research proof [14] that the characteristic of the gravity center does not behave the same as in obstacle planner space. Instead of representing the clusters by their gravity centers, the k-medoids algorithm chosen an actual object in the cluster as the clusters representative (medoid). Using the real object decreases the k-medoids sensitivity to outliers. This technique also guarantees that the center is accessible by all data objects within the same cluster. By comparing CLARA and CLARANS with PAM, CLARA first draws random samples of the data set and then do PAM on these samples. Unlike CLARA, CLARANS draws a random sample from all the neighbor nodes of the current node in the searching graph. Efficiency depends on the sample size and a good clustering based on samples will not necessarily represent a good clustering of the whole data. PAM is most accurate algorithm in partitioning based clustering algorithm because it's flexibility to check all nodes in each cell to determine the best location for base station. Therefore we were chosen as backbone of distance constraint clustering.

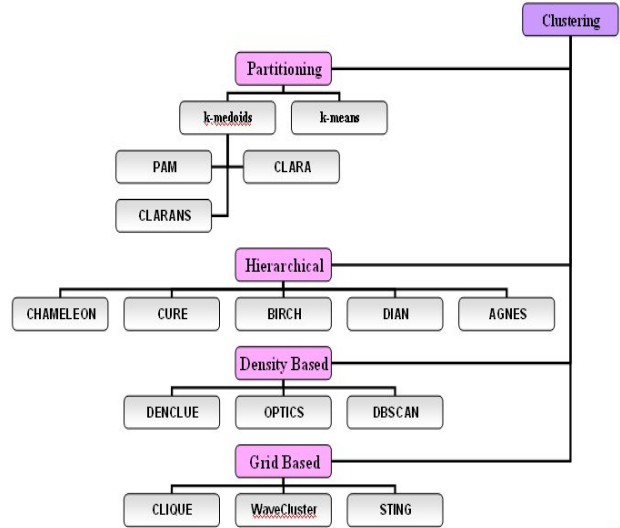

Fig. 2 Categorization of major clustering methods

The PAM (Partioning Around Medoids) algorithm, also called the K-medoids algorithm, represents a cluster by a medoid[16]. Initially, the number of desired clusters is input and a random set of k items is taken to be the set of medoids. Then at each step, all items from the input dataset that are not currently medoids are examined one by one to see if they should be medoids. That is, the algorithm determines whether there is an item that should replace one of the existing medoids. By looking at all pairs of medoids, non-medoids objects, the algorithm chooses the pair that improves the overall quality of the clustering the best and exchanges them. Quality here is measured by the sum of all distances from a non-medoid object to the medoid for the cluster it is in. A item is assigned to the cluster represented by the medoid to which it is closest (minimum distance or direct Euclidean distance between the customers and the center of the cluster they belong to).

The PAM algorithm [19] is shown in Fig. 3. We assume that $K_i$ is the cluster represented by medoid $t_i$. Suppose $t_i$ is a current medoid and we wish to determine whether it should be exchanged with a non-medoid $t_h$. we wish to do this swap only if the overall impact to the cost (sum of the distances to cluster medoids) represents an improvement. The total impact to quality by a medoid change $TC_{ih}$ is given by:

$$TC = \sum_{h=1}^{k} \sum_{n_i \in C_h} dis(n_h, n_i) \quad (1)$$

```
Algorithm PAM
Input:
D = {t_1, t_2, t_3, ..........., t_n} // set of elements
 A   // adjacency matrix showing distance between
    elements.
k   // Number of desired clusters.
Output:
K   // set of clusters.
PAM  Algorithm:
  Arbitrarily select k medoids from D;
 repeat
   For each t_h not a medoid do
    For each medoid t_i do
         Compute square error function TC_ih;
       Find i, h where TC_ih is the smallest;
       If  TC_ih  < current TC_ih then
         Replace medoid t_i with t_h;
   Until TC_ih  >= current TC_ih;
 For each t_i ∈ D do
 Assign t_i to K_j where dis(t_i , t_j ) is the smallest over all medoids;
```

Fig. 3 PAM algorithm

## 4. The M-PAM Algorithm

In a certain area contains number of subscribers distribution information, we need to determine the number of base stations and define their boundaries which satisfy good quality of service requirements with minimum cost.
The problem statement:-

- A set P data points {p1, p2,...., pn} in 2-D map, subscribers loads and communication constraints .
- Objective: Partition the city into k clusters C1, C2, .., Ck that satisfy clustering constraints (good quality of service) such that the cost function is minimized.
- Input: Set of n objects (map), set of intersection nodes (nodes contains number of subscribers).
- Output: number of clusters, Base Station locations, boundaries of each cluster.

The proposed algorithm contains three phases. Phase I is the pre-planning stage. We convert map from raster form to digital form that able to extract information from it, then we save digital map to database. Since PAM needs to determined number of clusters k in advance, we modified PAM by predicate the value of k. We used GSM technology to determine the number of cell needed by coverage planning and the number of cell needed by capacity planning, the initial k is the maximum of the two values Phase II is the main planning stage. In this phase we break down database to several clusters by applying clustering algorithm M-PAM  after knowing initial k, we determine the optimal location of base stations and its boundaries. The final step on this phase is calculating for each cluster the coverage and capacity plan. Phase III is the adjustment stage. If coverage or capacity plan of any cluster need more than one base station, applied one of the following methods. Method I: we increase number of clusters on the whole data, then go to the Phase II. Method II: we increase number of clusters on just the cluster that had a problem on its mobile constraints, then go to the Phase II.

I.     **Phase I : Pre-Planning**

This phase is divided into two steps. Step 1, convert map from raster form to digital form. Step 2, determine the initial number of clusters.

### - Map and their data entries

The maps used for planning are scanned images obtained by the user. It's need some preprocessing operations before it used as digital maps, we draw the streets and intersection nodes on the raster maps, the beginning and ending of each street are transformed into data nodes, defined by their coordinates. The streets themselves are transformed into links between data nodes. The subscriber's loads are considered to be the weights for each node. Fig. 4 show map transformation. For each intersection node and street the user can right click to input the characteristics of intersection node (no, name, capacity) or street (street number, street name,  street load).

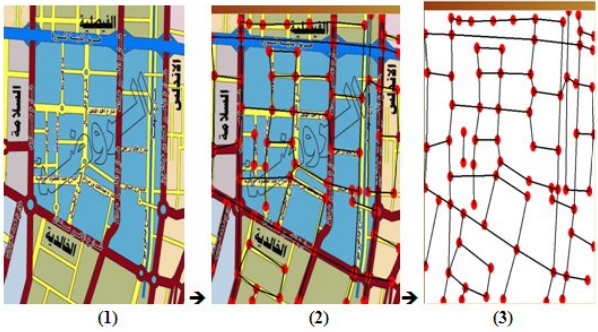

Fig. 4 map transforming
   (1) raster map
   (2) draw squares and streets on raster map
   (3) map after transforming to digital map

**- Determine Initial Number of Clusters**
In this work we used GSM technology of radio network planning to calculate number of cell need by coverage planning and calculate number of cell need by capacity planning for planned area.

Number of cell need by coverage planning = total area / area of the cell

Number of cell need by capacity planning = total number of subscribers / total subscribers per cell

Initial number of clusters k = the maximum of two values.

## II. Phase II : Main-Planning stage

In this phase, the goal is to split the entire database into clusters.

**- Partition Database**
After initial k is known, we used M-PAM algorithm to determine the optimal location of base station and its boundary of the served area for each cluster. The M-PAM algorithm is based mainly on the idea of the Partitioning Around Medoids (PAM). We imbedded the capacity and coverage algorithm to initiate the variable k. Fig. 5 shown the detail of M-PAM algorithm.

- Determined mobile constraints for Clusters
The next step, we used GSM technology of radio network planning to calculate number of cell needed by coverage planning and calculate number of cell needed by capacity for planned area. For each cluster we do the following.

Number of cell need by coverage planning = cluster area / area of the cell

Number of cell need by capacity planning= total number of subscribers per cluster / total subscribers per cell

Number of clusters K = the maximum of two values.

## III. Phase III : Adjustment stage

This phase, based on the maximum number of cluster (k) needed for each cluster which determined in previous section. The base station satisfy the constraints if and only if (k<=1) of each cluster. If any cluster not satisfies the requirements, we adding more clusters and redistribute nodes to close base station by Appling clustering algorithms again (M-PAM method I or M-PAM method II).

---

**Algorithm M-PAM**
**Input**:
  Set of *n* objects (maps), set of intersection nodes that contains number of subscribers.
  D = {$n_1$, $n_2$, $n_3$, ……….} // set of elements
**Output**:
  A partition of the d objects into k clusters and clusters centre, $m_1$, $m_2$...... $m_k$
**Algorithm:**
  Compute k   ( Number of desired clusters)
    Call  coverage algorithm     ' coverage of BS
    Call  capacity algorithm     ' capacity of BS
  number- of- cells1= Surface of area to be plan / coverage of BS
  number- of- cells2= number of subscribers of area to be plan /    capacity of BS
  k = max(number-for-cells1 , number-for-cells2)
Label 1   Arbitrarily select k medoids from D;
    repeat
      For each $n_i$ not a medoid do
        For each medoid $n_h$ do
          Compute function $TC_{ih}$;
          Find i, h where $TC_{ih}$ is the smallest;
          If  $TC_{ih}$ is improved then
            Replace medoid $n_h$ with $n_i$;
    Until $TC_{ih}$ is not improved;
    For each $n_h \in$ D do
      Assign $n_h$ to $n_j$ where dis($n_h$ , $n_j$ ) is the smallest over all medoids
    For {I = 1 to k }   /* k = number of cluster
      Call  caverage algorithm
      Call  capacity algorithm
  number- of- cells1= Surface of cluster / coverage of BS
    number- of- cells2= number of subscribers of cluster / capacity of BS
    If (number-for-cells1  > 1 or number-for-cells2 >1)
      Then k = k + 1  go to label 1       'for method I'
      'for method II, do the followings:
      k= k+1 then       'on cluster had a problem
        Apply cluster algorithm on cluster($d_i$)  go to label 1

Fig. 5  M-PAM Algorithm

## 5. Case Study

By applying M-PAM algorithm method I and method II to different datasets Table 2 and different cell range we obtain the results shown in figure 6 – figure 7. The experiments show that method II does not have any better solution with small datasets (like dataset I and dataset II). But for large datasets with large number of subscribers are deployed in different area, it produces a better solution and cost minimization. Because we applied the clustering algorithm only on the clusters that had a problem on its mobile requirements whether coverage or capacity or even both instead of clustering a whole planned area by adding more clusters and redistribute nodes by using clustering algorithms which sometimes repeat the problem that depending on the redistributing process.

Figure 8 and 11 show the results of applying M-PAM algorithm method I and method II to the same numbers of nodes but when increase density (numbers of subscribers).

Table 2 Data base entries for comparison

| Data set / Number of Nodes | Coverage Area [m$^2$] | Number of Subscribers |
|---|---|---|
| Data set I = 50 | 230850 | 3139 |
| Data set II = 70 | 335478 | 3500 |
| Data set III = 101 | 337800 | 4000 |
| Data set IV = 150 | 345663 | 4488 |
| Data set V = 300 | 394284 | 10159 |

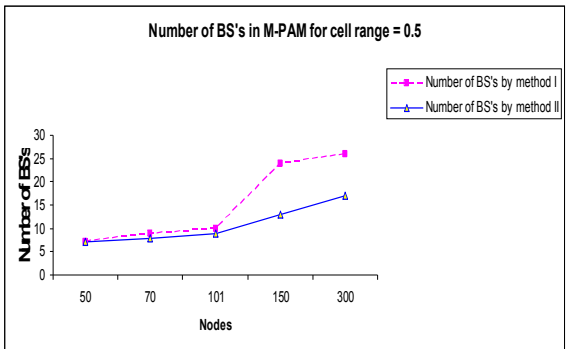
Fig. 6 Number of BS's in M-PAM (method I Vs. method II) for cell=0.5

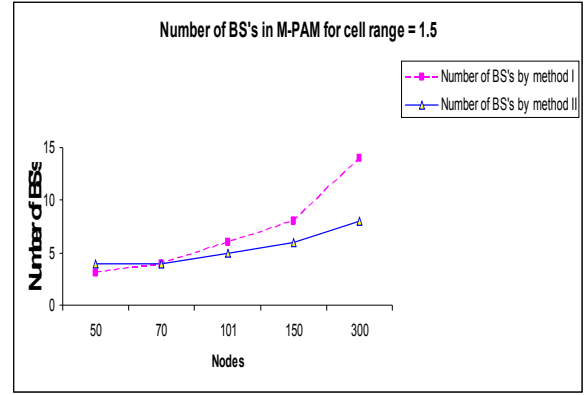
Fig. 7 Number of BS's in M-PAM (method I Vs. method II) for cell=1.5

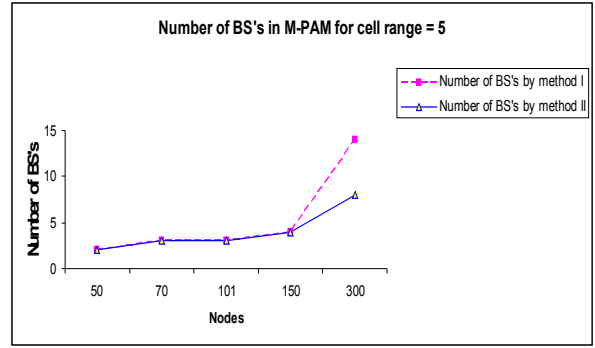
Fig. 8 Number of BS's in M-PAM (method I Vs. method II) for cell=5

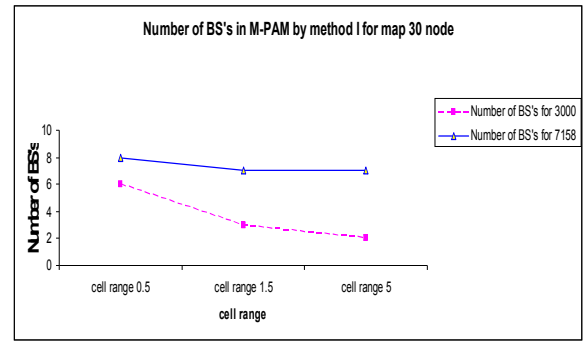
Fig. 9 Number of BS's in M-PAM by method I for 30 node

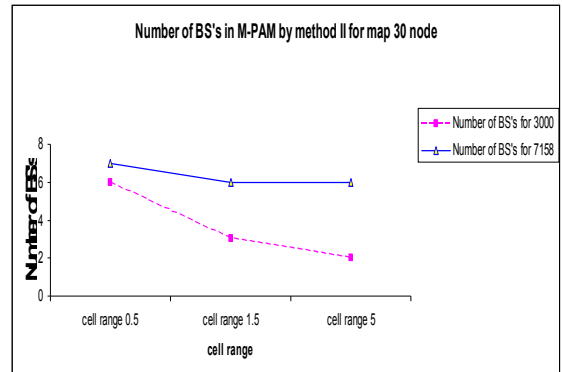
Fig. 10 Number of BS's in M-PAM by method II for 30 node

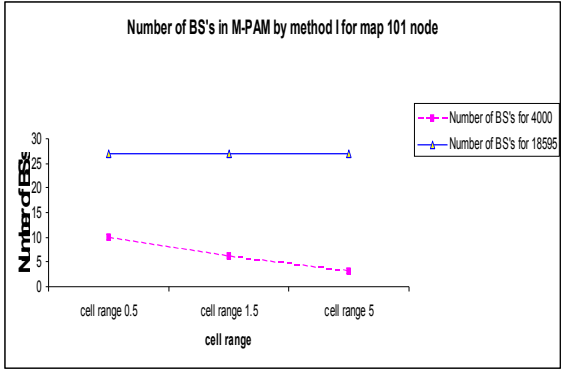

Figure 11 Number of BS's in M-PAM by method I for 101 nodes

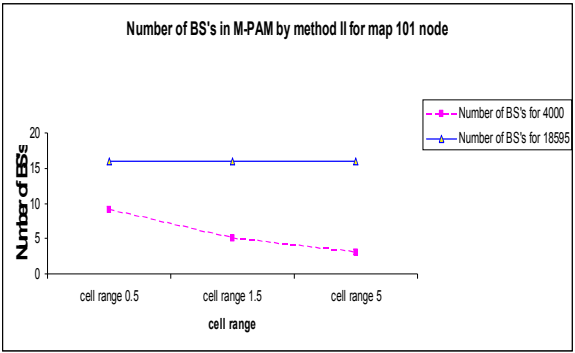

Figure 12 Number of BS's in M-PAM by method II for 101 nodes

With 30 nodes and 3000 subscribers and with 101 nodes and 4000 subscribers, since the number of subscribers is small, the number of base stations had different values with different cells range. With 30 nodes and 7158 subscribers and with 101 nodes and 18595 subscribers, since the number of subscribers is big, therefore the problem in these maps is in their capacity. The number of base stations had not based effected with different cells ranges. But also method II obtain the minimum numbers of Base station.

Fig. 13 to 15 show a comparison between the proposed algorithms M-PAM (method I and method II) and the original PAM algorithm for cell range=0.5, 1.5 and 5 km with respect to run time. The big values of run time in PAM are due to the estimated value of K which is factor of the expertise of the user of the program which may be increase or decrease accordingly.

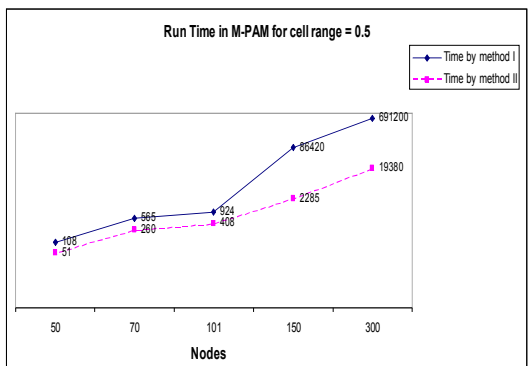

Fig. 13 Run time in M-PAM (method I Vs. method II) for cell=0.5

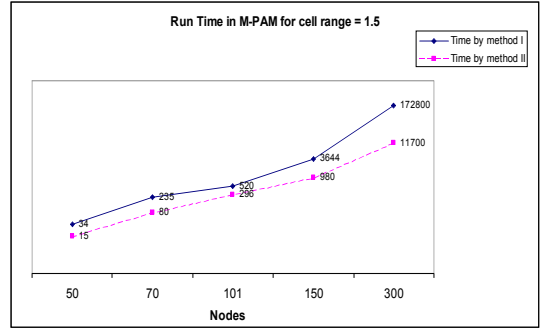

Fig. 14 Run time in M-PAM (method I Vs. method II) for cell=1.5

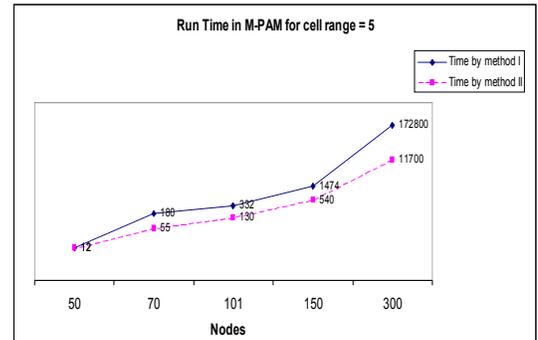

Fig. 15 Run time in M-PAM (method I Vs. method II) for cell=5

## 6. Comparison between Proposed Method and Other Methods

Table 3 described the different comparison between the proposed method and other methods using in mobile network planning. There are two methods that are frequently used here: Tabu search and Genetic Algorithms.
Tabu Search: Tabu search is defined as follows: The already selected base station positions in the last K iterations will be considered "tabu", and so will not be taken into consideration for the generation of new neighbors. Usually K is chosen to be 1. The size of candidates for each iteration can be reset, and the larger the size of candidates, the better the final result, but also it takes longer for the simulations to converge. The best value for the candidate size can be found to be equal to 10 in the literature.
Genetic Algorithm: A genetic algorithm behavior mimics the evolution of simple, single celled organisms. It is particularly useful in situations where the solution space to be searched is huge, making sequential search computationally expensive and time consuming.

Table 3 shows the comparison between relative works. Tabu Search and Genetic algorithm, needed a huge numbers of estimated input parameters which can be affected to the results.

## 7. Conclusion and Future Work

In this paper the proposed algorithm M-PAM, which modifies clustering technique PAM to solve the problem of mobile network planning, is presented. These algorithms are medoid clustering algorithm. PAM needs to specify number of clusters (k) before starting to search for the best locations of base stations. The M-PAM algorithm uses the radio network planning to determine k. We calculate for each cluster its coverage and capacity and determine if they satisfy the mobile requirements, if not we will increase (k) and reapply algorithms depending on two methods for clustering. Implementation of this algorithm to a real case study is presented. Experimental results and analysis indicate that the M-PAM algorithm when applying method two is effective in case of heavy load distribution, and leads to minimum number of base stations, which directly affected onto the cost of planning the network. It is expected that by applying this system to a number of areas belonging to different countries with different sizes, one can verify its capabilities more universally. The next generation mobile communication system is desired to transmit multimedia information at multi-rate. Therefore, we can implement the UMTS technology instead of GSM technology by modify only the coverage and capacity planning algorithms.

**Lamiaa Fattouh Ibrahim** is a Faculty Member in College of Computing and Information Technology-King AbdulAziz University in Jeddah, Faculty Member in Institute of Statistical Studies and Research - Cairo University. She holds a Doctor of Philosophy

P.h.D. from Cairo University, Faculty of Engineering, 1999, Master from Ain Shams University, Faculty of Engineering, Computer & Systems Engineering Department 1993, Master from Ecole National Superieur de Telecommunication, ENST Paris 1987 and B.S.c from Ain Shams University, Faculty of Engineering, Computer & Automatic control Department 1984. Over 26 years of experience in the fields of network engineering and artificial Intelligent, with focus on applying knowledge base and data mining techniques in wire and wireless network planning. She has published many papers in many journals and international conferences on areas network, data mining, wire and Mobile network planning. She is a member of the editorial board of the International Journal on Advances in Telecommunications, IARIA, and Journal of Computer Science, Science Publications. She is reviewer in Journal of Network and Computer Applications, Elsevier.

**Manal Hamed Al Harbi** Lecture College of Education, UMM AL-QURA University, Macca, Saudi Arabia.

Table 3 Relative Work

| Algorithm | Input parameters | Results | Location of BS'S | Constraints | Type of Distance |
|---|---|---|---|---|---|
| Genetic Algorithm | -Data points<br>-Population size<br>-Initial probability<br>-Mutation probability<br>-Crossover probability<br>-Number of iteration<br>-Selection pressure | # of clusters | Optimal placement | Fitness Function | Euclidean distance |
| Tabu Search | -Data points<br>-Init probability<br>-Generation probability<br>-Recency factor<br>-Frequency factor<br>-Number of iteration<br>-Number of neighbors | # of clusters | Optimal placement | Tabu List | Euclidean distance |
| PAM | -Data points<br>-K | -Clusters medoid<br>-# of clusters | Medoids | $TC_{ih} = \sum C_{ih}$ | Euclidean distance |
| M-PAM | -Data points | -Clusters medoid<br>-# of clusters | Medoids | $TC_{ih} = \sum C_{ih}$ | Euclidean distance |